\documentclass[10pt,twocolumn,letterpaper]{article}

\usepackage{cvpr}
\usepackage{times}
\usepackage{epsfig}
\usepackage{graphicx}
\usepackage{amsmath}
\usepackage{amssymb}

\usepackage{multirow}
\usepackage{color}
\usepackage{algorithm}
\usepackage{algorithmic}
\usepackage{array}
\usepackage{stfloats}
\usepackage{url}
\usepackage{ragged2e}
\usepackage{cite}
\usepackage{cases}
\usepackage{subfig}
\usepackage{setspace}

\graphicspath{{./figurepdf/}}
\DeclareGraphicsExtensions{.pdf}

\definecolor{americanrose}{rgb}{1.0, 0.01, 0.24}

\makeatletter
\def\hlinew#1{%
  \noalign{\ifnum0=`}\fi\hrule \@height #1 \futurelet
   \reserved@a\@xhline}
\makeatother%

\usepackage[breaklinks=true,bookmarks=false]{hyperref}

\cvprfinalcopy % *** Uncomment this line for the final submission

\def\cvprPaperID{906} % *** Enter the CVPR Paper ID here

\ifcvprfinal\fi

\begin{document}

%%%%%%%%% TITLE
\title{C-MIL: Continuation Multiple Instance Learning for \\Weakly  Supervised Object Detection}

% \author{Fang Wan\\
% \and

\author{
Fang Wan$^\dag$, Chang Liu$^\dag$, Wei Ke$^\dag$, Xiangyang Ji$^\ddag$, Jianbin Jiao$^\dag$ and Qixiang Ye$^{\dag\S*}$\\
\\
$^\dag$University of Chinese Academy of Sciences, Beijing, China\\
$^\ddag$Tsinghua University, Beijing, China.   $^\S$Peng Cheng Laboratory, Shenzhen, China\\
{\tt\small \{wanfang13,liuchang615,kewei11\}@mails.ucas.ac.cn, xyji@tsinghua.edu.cn}\\
{\tt\small \{jiaojb,qxye\}@ucas.ac.cn}
}

\maketitle
%\thispagestyle{empty}

%%%%%%%%% ABSTRACT
\begin{abstract}
   Weakly supervised object detection (WSOD) is a challenging task when provided with image category supervision but required to simultaneously learn object locations and object detectors. Many WSOD approaches adopt multiple instance learning (MIL) and have non-convex loss functions which are prone to get stuck into local minima (falsely localize object parts) while missing full object extent during training. In this paper, we introduce a continuation optimization method into MIL and thereby creating continuation multiple instance learning (C-MIL), with the intention of alleviating the non-convexity problem in a systematic way. We partition instances into spatially related and class related subsets, and approximate the original loss function with a series of smoothed loss functions defined within the subsets. Optimizing smoothed loss functions prevents the training procedure falling prematurely into local minima and facilitates the discovery of Stable Semantic Extremal Regions (SSERs) which indicate full object extent. On the PASCAL VOC 2007 and 2012
   %, and MS-COCO 2014
   datasets, C-MIL improves the state-of-the-art of weakly supervised object detection and weakly supervised object localization with large margins {\let\thefootnote\relax\footnotetext{\vspace{-0.8em}\par\setlength{\parindent}{1.4em}$^*$Corresponding author.}}\footnote{The code for CMIL is available at github.com/Winfrand/C-MIL.}.
   %but required to learn, at the same time, object locations and object detectors.
   %Many WSOD approaches that adopt have non-convex objective functions and are prone to get stuck into local minima (falsely localize object parts) while missing true object extent during training.  In this paper, we introduce a continuation optimization method into MIL and update it to continuation multiple instance learning (C-MIL), with the target of alleviating the non-convexity problem in a principled way. We partitioned instances into spatially related and class-correlated subsets, and approximate the original objective function with a series of smoothed functions defined on the subsets.
   %Optimizing the smoothed functions prevents the training procedure prematurely falling into local minima and facilitates discovering Stable Semantic Extremal Regions (SSERs) indicating true object extent. On the Pascal VOC 2007, 2012, and MSCOCO 2014 datasets, C-MIL improves the state-of-the-arts of weakly supervised object detection and localization with {\color{blue}large} margins \footnote{The code for C-MIL is available at XXXXXXXXXXXXXXX}.
\end{abstract}

%%%%%%%%% BODY TEXT
\section{Introduction}

\thispagestyle{empty}

\begin{figure*}[h]
    \begin{center}
        \includegraphics[width=0.95\linewidth]{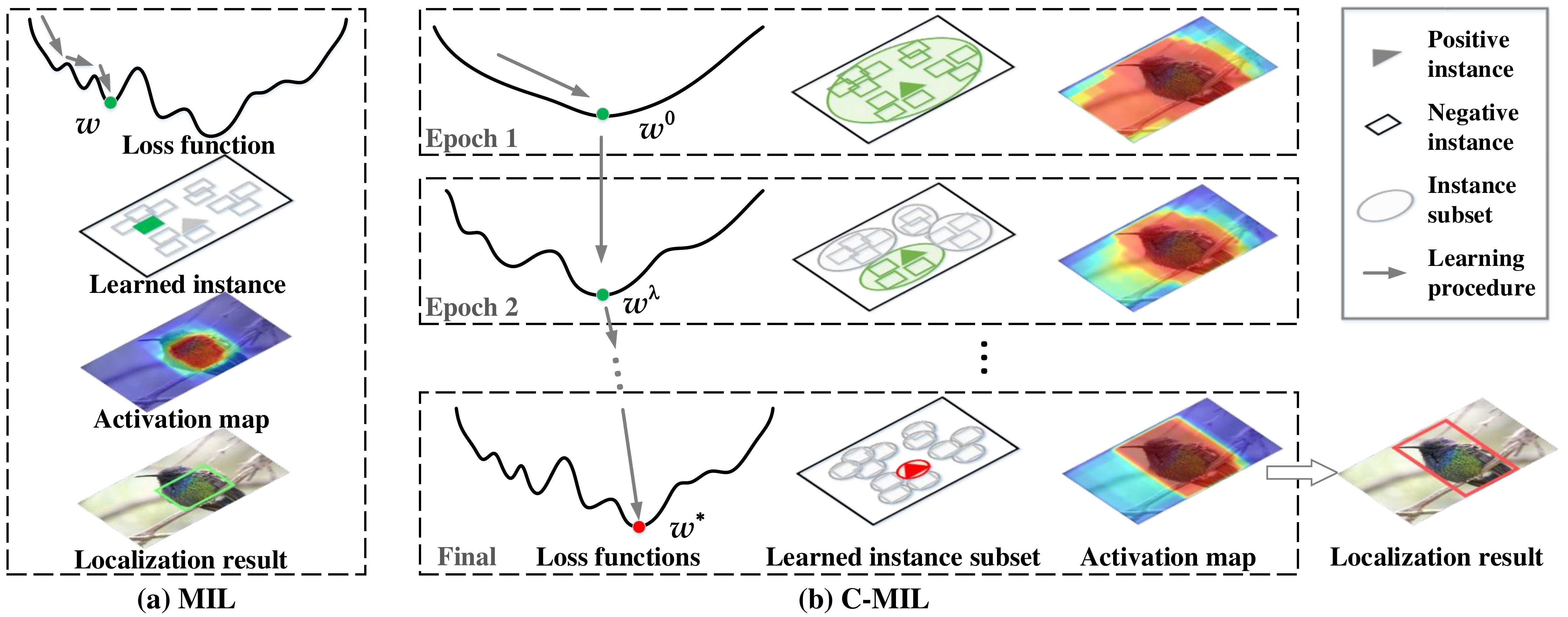}
        % \subfloat[MIL]{  \includegraphics[height=5.2cm]{Fig-Motivation-a}}
        % \subfloat[C-MIL]{\includegraphics[height=5.2cm]{Fig-Motivation-b}}
    \end{center}
    \vspace{-0.3cm}
    \caption{
    Comparison of MIL-based and C-MIL-based WSOD approaches. Due to the non-convex loss function MIL often falls into local minima and falsely localizes an object part. By introducing the continuation optimization with a series of smoothed loss functions, C-MIL alleviates the non-convexity problem and localizes full object extent. (Best viewed in color)
    }

    \label{fig_motivation}
    \vspace{-0.1cm}
\end{figure*}

Weakly supervised object detection (WSOD) is rapidly gaining attention in computer vision area. WSOD approaches only require image category annotations indicating the presence or absence of a category of objects in images, significantly reducing human involvement by omitting labor-intensive bounding-box annotations \cite{bilen2014, song2014pattern, Ye2017SelLearning, wang2014LCL, Song2014On, Siva2011Weakly}.

Despite extensive research over the past five years, WSOD remains an open problem, as indicated by the large performance gap ($\sim$ 20\%) between WSOD \cite{Tang2017OICR,wan2018melm,Wei2018TS2C} and fully supervised detection approaches \cite{ren2015faster-rcnn,girshick2015fast-rcnn} on the PASCAL VOC detection benchmark \cite{everingham2010pascal}.

Combined with deep neural networks, MIL has been the main WSOD method \cite{Bilen2016Weakly,Tang2017OICR}. However, it is observed that the model is prone to activate object parts instead of full object extent, particularly during the early learning epochs, Fig.\ \ref{fig_motivation}(a).
This phenomenon arises from non-convexity of the objective/loss functions. Optimizing such functions can get stuck into local minima, $i.e.,$ selecting most discriminative regions (instances) for image classification while ignoring full object extent \cite{bilen2015,wan2018melm}.

Researchers have alleviated this problem by using spatial regularization \cite{Bilen2016Weakly, Diba2017WCCN, wan2018melm}, context information \cite{Kantorov2016ContextLocNet,Wei2018TS2C}, and progressive refinement \cite{Li2016Weakly, Diba2017WCCN,Ye2017SelLearning,Tang2017OICR,wan2018melm}. Despite their advances, the local minimum problem remains unsolved from an optimization perspective.

In this paper, we introduce the continuation method \cite{Allgower1990Numerical}, which addresses a complex optimization problem
by smoothing the loss function and turning it into multiple easier sub-problems, into multiple instance learning and thereby creating continuation multiple instance learning (C-MIL), with the purpose of alleviating the non-convexity problem in a systematic manner. C-MIL treats images as bags and image regions generated by an object proposal method \cite {uijlings2013selective,Zitnick2014Edge} as instances.
%to train instance classifiers (object detectors).
During training, unlike the conventional MIL that pursues the most discriminative instances, C-MIL learns instance subsets, where the instances are spatially related, $i.e.,$ overlapping with each other, and class related, $i.e.,$ having similar object class scores. Instance subsets with proper continuation parameters are capable of collecting object parts to fine-tune the network, and activate Stable Semantic Extremal Regions (SSERs) indicating full object extent, Fig.\ \ref{fig_motivation}(b).

Instance subsets are partitioned according to continuation parameters. With the smallest parameter, an image is partitioned into a single subset which contains all instances while the loss function of C-MIL is equal to that of image classification which is convex. With the largest parameter, each instance is defined as a subset, and the loss function degenerates to that of MIL. During training, the continuation parameter gradually dwindles the subset from the maximum set (with all instances) to the minimum sets (with a single instance). In this way, we construct a series of functions which are easier to optimize to approximate the original loss function, Fig.\ \ref{fig_motivation}(b). With end-to-end training, the most discriminative subset in each image is discovered, and subsets/instances which lack discriminative information are suppressed.
%The instance numbers in the subsets are determined by a continuation parameter. While a subset covering all instances in an image, the loss function of C-MIL is equal to that of image classification and is convex. While a subset containing a single instance, the loss function degenerates to that of MIL. During training, the continuation parameter is gradually changed to dwindle each subset from the maximum set (with all instances in the image) to the minimum set (with a single instance). In this way, we construct multiple smoothed functions those are easy to be optimized to approximate the non-convex loss function, Fig.\ \ref{fig_motivation}. With end-to-end training in a deep learning framework, discriminative instance subsets containing objects/parts are discovered, and the subsets/instances that lack discriminative information are suppressed.

The contributions of this paper include:
%This study contributes to the field in the following ways:

(1)	A novel C-MIL approach which uses a series of smoothed loss functions to approximate the original loss function, alleviating the non-convexity problem in multiple instance learning.

(2)	A parametric strategy for instance subset partition, which is combined with a deep neural network to activate full object extent.

(3)	New state-of-the-art performance of weakly supervised detection and localization on commonly used object detection benchmarks.

\section{Related Work}

%WSOD problems are often solved with a pipelined approach, i.e., an object proposal method is first applied to decompose images into object instances, with multiple instance learning is used to iteratively perform instance selection and detector estimation. With the widespread acceptance of deep learning, pipelined approaches have been evolving into multiple instance learning networks.
For many branches of WSOD methods, \cite{wang2014LCL,wang2015LCL,Ye2017SelLearning,Song2014On, bilen2015}, we mainly review MIL-based approaches. We also review the continuation optimization and smoothing methods for the non-convex optimization.

\subsection{Weakly Supervised Methods}

\textbf{MIL.} As the major line of WSOD method, MIL treats each training image as a ``bag'' and iteratively selects high-scored instances from each bag when learning detectors. It works in a similar way to the Expectation-Maximization algorithm estimating instances and detectors simultaneously. Nevertheless, such an algorithm is frequently puzzled by local minima caused by non-convex loss functions, particularly, when the solution space is large \cite{bilen2015, wan2018melm}.

To alleviate the non-convexity problem, clustering was used as a pre-processing step to facilitate instance selection considering that a class of instances often shape a single compact cluster \cite{Song2014On,wang2015LCL, bilen2015}.
%while the negative instances form multiple diffuse clusters
%In \cite {wang2015LCL}, Wang et al. calculated clusters of object instances using probabilistic latent Semantic Analysis (PLSA) on the regions of positive samples, and employed these clusters to determine positive sub-categories. in [3] and [], Bilen and Song leveraged clustering to initialize latent variables, i.e., object regions, part configurations and sub-categories, and learn object detectors based on the initialization.
A bag splitting strategy was proposed to reduce the solution space during the optimization procedure of MILinear \cite{Ren2016Weakly}. The multi-fold MIL \cite{cinbis2014multi-fold-C,cinbis2015multi-fold-J} with training set partition and cross validation was proposed to realize multi-start point optimization.
%Although effective in reducing the solution spaces, these approaches usually lack a principled way to alleviate the non-convexity problem.
%The disadvantage lies in that a positive cluster could incorporate significant noise if the objects are surrounded by clutter backgrounds.
%Hoffman et al. train detectors with weakly annotations while transferring representations from extra object classes using full supervision (bounding-box annotation) and joint optimization.

\textbf{MIL Networks.}
%MIL has been updated to deep multiple instance learning networks, where the convolutional filters behave as detectors to activate regions of interest on the deep feature maps. Li et al. have adopted progressive optimization to step-wisely learn object information from image class label. Tang et al. propose to refine instance classifiers online by propagating instance labels to spatially overlapped instances. Diba et al. propose weakly supervised cascaded convolutional networks (WCCN) with multiple learning stages. Recent approaches have introduced the regularizers of entropy, object count, context, and multi-evidence to the MIL framework to deal with the non-convexity problem. Wan et al. have introduce entropy models to avoid the localization randomness. Gao et al. used extra supervision of object count for each image. Wei et al. and Kantorov et al. have improved the localization accuracy by using object context information. Ge et al. have used multi-evidence of objects in weakly supervised learning. Tang et al. and Shen et al. have improved the detection speed by training a RPN or SSD. Zhang et al. and Zhang et al. utilized the size and spatial prior of instances by observing the off-the-shelf WSOD methods. However, these methods always try to localize the object directly through the image label, which makes them susceptible to the local minima produced by the image classification framework.
MIL has been updated to MIL networks \cite{Bilen2016Weakly}, where convolutional filters behave as detectors which activate regions of interest on the feature maps. However, loss functions of MIL networks remain non-convex and thus suffer from local minima. To alleviate this problem, researchers introduced spatial regularization \cite{Bilen2016Weakly, Diba2017WCCN, wan2018melm}, context information \cite{Kantorov2016ContextLocNet,Wei2018TS2C}, and progressive optimization \cite{Li2016Weakly, Diba2017WCCN,Ye2017SelLearning,Tang2017OICR,wan2018melm} into the MIL networks.

In \cite{Diba2017WCCN}, object segmentation was used as the regulariser and optimized with instance selection in two learning stages within cascaded convolutional networks. In \cite{wan2018melm}, a clique-based min-entropy model was proposed as the regularizer to alleviate localization randomness during learning instances. In \cite{gao2018cwsl}, the per-class object count was leveraged to address failure cases about one detected box containing multiple instances. In \cite{Kantorov2016ContextLocNet,Wei2018TS2C}, context models were designed to learn instances while being both supported by and standing out from surrounding regions.
%to be supported by their surrounding regions as well as outstanding from these surrounding regions.

Existing methods often use high-quality regions (instances) as pseudo ground-truth
to progressively refine the classifier \cite{Li2016Weakly,Diba2017WCCN,Tang2017OICR,wan2018melm}. In \cite{Tang2017OICR}, an online instance classifier refinement algorithm was integrated with the MIL network.
%with an instance classifier refinement in the deep learning framework.
In \cite{wan2018melm}, a recurrent learning algorithm was proposed to integrate image classification with object detection, and then to progressively optimize the classifiers and detectors.

Existing strategies using spatial regularization, context information, and progressive refinement are effective at improving WSOD. Nevertheless, there still lacks a principled and systematic way to alleviate the local minimum problem from the perspective of optimization.
%principled ways to alleviate the local minimum problem from the perspective of optimization.
%In this paper, we propose a continuation optimization method \cite{Allgower1990Numerical,Gulcehre2017Mollifying}.

\subsection {Non-convex optimization}
\textbf{Continuation methods.} Continuation methods \cite{richter1983continuation, allgower1994numerical} address a complex optimization problem by smoothing the loss function, turning it into multiple sub-problems which are easier to optimize. By tuning continuation parameters, it incorporates a sequence of sub-problems which converge to the optimization problem of interest. These methods have been successful in tackling optimization problems involving non-convex loss functions with multiple local minima.
%and possibly points of non-differentiability.
In machine learning, curriculum learning \cite{bengio2009curriculum} was inspired by this principle to define a sequence of gradually increasing difficulty training tasks (or training distributions) which converge to the task of interest. Gradient-based optimization over a sequence of mollified loss functions has been shown converging to stronger global minima \cite{chen2012smoothing}.

\textbf{Smoothing.} Smoothing is an important technique in optimization \cite{beck2012smoothing} and has been applied in deep neural networks.
In \cite{zheng2015improving} and \cite{clevert2015fast}, a method which modified the non-smooth ReLU activation to improve training was proposed. In \cite{Gulcehre2017Mollifying}, ``mollifiers" were introduced to smooth the loss function by gradually increasing the difficulty of the optimization problem. In \cite{chaudhari2017entropy}, entropy was added to the loss function to promote solutions by reducing randomness.
%In particular, Zheng et al. \cite{zheng2015improving} and Clevert et al. \cite{clevert2015fast} proposed modifying the non-smooth ReLU activation to improve the training. Gulcehre et al. \cite{Gulcehre2017Mollifying} introduced ``mollifiers" to smooth the objective function by gradually increasing the difficulty of the optimization problem. Chaudhari et al. \cite{chaudhari2017entropy} added a local entropy term to the loss to promote solutions with high local entropy.

In this study, we implement continuation optimization by specifying a series of smoothed loss functions for a MIL network over spatially related and class related instance subsets, and target at alleviating the local minimum problem and learning full object extent.
%in a deep learning framework by . We target at exploring a principled way to alleviate the local minimum problem of MIL networks.

%------------------------------------------------------------------------

\begin{figure*}[!t]
    \begin{center}
       \includegraphics[width=0.87\linewidth]{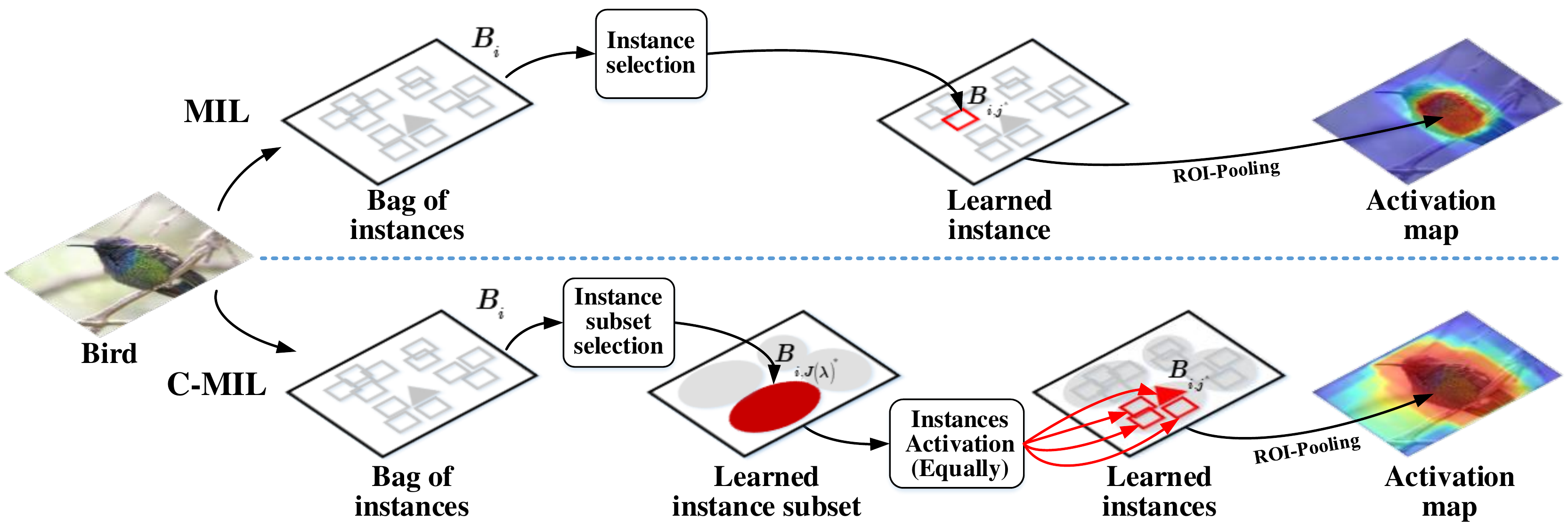}
    \end{center}
    \vspace{-0.6cm}
    \caption{
    {\color{black} Comparison of the instance selection strategies of MIL and C-MIL. MIL tends to select the most discriminative instance and activate the object part. In contrast, C-MIL selects the most discriminative instance subset. The instances in the subset are activated equally during back-propagation and thus the object extent is activated. (Best viewed in color)}
    }
    \label{fig_instance_set}
    \vspace{-0.5em}
\end{figure*}

\section{Methodology}

C-MIL treats images as bags and image regions generated by an object proposal method \cite {uijlings2013selective,Zitnick2014Edge} as instances. The goal is to train instance classifiers (detectors) while solely the bag labels are available.
In Fig.\ \ref{fig_instance_set}, $B_i \in {\cal B}$ denotes the $i^{th}$ bag (image) and ${\cal B}$ denotes all bags (training images). $y_i \in {\cal Y}$ where ${{\cal Y}=\{1,-1\} }$ denotes the label of bag $B_i$ indicating the bag contains positive instances ($i.e.$, objects with positive class) or not. $y_i=1$ indicates a positive bag (image) that contains at least one positive instance, while $y_i=-1$ indicates a negative bag where all instances are negative.
Let $B_{i,j}$ and $y_{i,j}$ denote instances and instance labels in bag $B_i$, where $j \in \{1,2,...,N\}$ and $N$ the number of instances. $w$ denotes network parameters to be learned.

%With these definitions, we first revise the MIL method for WSOD and then analyze its convexity. We then present the C-MIL approach and its implementation with a convolutional neural network (CNN).

\subsection{MIL Revisit}

With above definitions, an MIL method \cite{andrews2002MIL,cinbis2014multi-fold-C,cinbis2015multi-fold-J} can be separated into two alternative steps: instance selection and detector estimation. In the instance selection step,
%an instance selector $f(B_{i,j}, w_f)$ is trained to mine the instance $B_{i,j^*}$ which is of highest score to be a positive instance (object), as
an instance selector $f(B_{i,j}, w_f)$, which computes the object score of each instance, is used to mine a positive instance (object) from $B_i$.
\begin{equation}
    B_{i,j^*} = \mathop {\arg \max }\nolimits_{{j}} f\left( {{B_{i,j}},{w_f}} \right),
\end{equation}
where $w_f$ indicates the parameters of the instance selector and $j^*$ the index of the selected instance of the highest score. With selected instances, a detector $g_z(B_{ij}, w_g)$ with parameter $w_g$ is trained, where $z\in{\cal Y}$. $w_f$ and $w_g$ respectively denote parameters for the instance selector and detector.

%In conventional MIL methods, the instance selector $f(B_{i,j}, w_f)$ and object detector $g(B_{i,j}, w_g)$ share the same model architecture and parameters, and they are often performed alternatively.
% $\frac{1}{2}{\left\| w \right\|^2}$
In MIL networks \cite{Bilen2016Weakly,Kantorov2016ContextLocNet,Tang2017OICR}, the two alternative steps are integrated and $f(B_{i,j}, w_f)$ and $g_z(B_{i,j}, w_g)$ are jointly optimized with loss functions on training images $\cal B$, as
\begin{equation}\label{eq_deep_MIL}
    {\cal F}({\cal B},w) = \sum\nolimits_i {{{\cal F}_f}\left( {{B_i},{w_f}} \right) + {{\cal F}_g}\left( {{B_i},B_{i,j^*},{w_g}} \right)},
\end{equation}
where the first term, loss of instance selection, is defined as
\begin{equation}\label{eq_deep_MIL_RegionMiner}
    {{\cal F}_f}\left( {{B_i},{w_f}} \right) = \max ( {0,1 - {y_i}\mathop {\max }\nolimits_j f\left( {{B_{i,j}},{w_f}} \right)}),
\end{equation}
which is the standard hinge loss. The second term, loss of detector estimation, is defined as
%\begin{equation}\label{eq_deep_MIL_ObjectDetector}
%    {{\cal F}_g}\left( {B_i,B_{i,j}^*,{w_f}}\right) = -\sum\limits_j {{1_{\left[ {{B_{i,j}},B_{i,j}^*} %\right]}}\log g\left( {{B_{i,j}},{w_g}} \right)},
%\end{equation}
%where $1_{\left[ {{B_{i,j}},B_{i,j}^*} \right]}$ is an indicator function. {\color{blue} If $B_{i,j}$ and  %$B_{i,j}^*$ belong to the same instance subset, it equals to 1; otherwise 0. $g(B_{i,j}, w_g)$ is in [0, 1].}
\begin{equation}\label{eq_deep_MIL_ObjectDetector}
    % \small %normalsize > small > footnotesize > scriptsize
    % {{\cal F}_g}\left( {B_i,B_{i,j^*},{w_g}}\right) = -\log g\left(B_{i,j^*},{w_g} \right).
    {{\cal F}_g}\left( {B_i,B_{i,j^*},{w_g}}\right) = -\sum\nolimits_z\sum\nolimits_j {{\delta_{z,y_{i,j}}}\log g_z\left( {{B_{i,j}},{w_g}} \right)},
\end{equation}
where $y_{i,j}$ is defined following the VOC metric \cite{everingham2010pascal} as
\begin{equation}\label{eq_label_assign}
    {y_{i,j}} = \left\{ {\begin{array}{*{20}{c}}
{+1,if\;IoU\left( {{B_{i,j}},B_{i,j^*}} \right) \ge 0.5}\\
{-1,if\;IoU\left( {{B_{i,j}},B_{i,j^*}} \right) < 0.5}
\end{array}} \right..
\end{equation}
$\delta_{a,b}$ is the Kronecker function which is defined as: $\delta_{a,b}=1\;if\;a=b$, and 0 otherwise.

\subsection{Convexity Analysis} %\textbf{\color{red}ONLY convexity analysis, NO Smoothness analysis}
Recall that the maximum of a set of convex functions is convex. When $y_i=-1$, Eq. \ref{eq_deep_MIL_RegionMiner} is convex, but when $y_i=1$, it is non-convex. The loss function (Eq. \ref{eq_deep_MIL}) of the MIL network is therefore non-convex as its first term (Eq. \ref{eq_deep_MIL_RegionMiner}) is non-convex,
%This is the primary problem that WSOD methods suffer from local minimum problem when selecting instance from $positive$ bags.
%Although the second term of Eq. \ref{eq_deep_MIL}, $i.e.$, the detector estimation loss, is convex, the convexity of Eq. \ref{eq_deep_MIL} is determined by the non-convex term Eq. \ref{eq_deep_MIL_RegionMiner}.
and it may have many local minima when provided with bags of numerous instances. Once false positives are mined by the instance selector, the detector will be misled by them, particularly in the early training epochs.

With above analysis, it is concluded that the following two problems remain to be elaborated: 1) How to optimize the non-convex function, and 2) How to perform instance selection in the early training stages when the instance selector is not well trained.

\subsection{Continuation MIL}
We propose a new optimization method, called Continuation Multiple Instance Learning (C-MIL), and target at solving the above two problems. Instead of introducing regularizers into the loss functions, we directly focus on them from an optimization perspective, by partitioning instances in a bag into subsets and manipulating the non-convexity or smoothness of the loss function defined by Eq. \ref{eq_deep_MIL_RegionMiner}.
%{\color {red}It is known that if a bag contains a single instance, the maximum term in brackets of Eq. \ref{eq_deep_MIL_RegionMiner} is linear. Eq. \ref{eq_deep_MIL_RegionMiner} is then convex when $y=1$, as a linear function is convex and the linear function multiplied with -1 remains linear. Based on this property, we propose using continuation method to optimize the instance selection and classifier estimation for alleviating the problems in the conventional deep MIL framework.}
%If we replace the term $\mathop {\max }\nolimits_j f\left( {{B_{i,j}},{w_f}} \right)$ of Eq. \ref{eq_deep_MIL_RegionMiner} by $\sum\nolimits_j {f\left( {{B_{i,j}},{w_f}} \right)}$, it changes to be convex. With $\sum\nolimits_j {f\left( {{B_{i,j}},{w_f}} \right)}$, all the instances in a bag are treated equally and . Based on this property, we propose using continuation method to optimize the instance selection and classifier estimation for alleviating the non-convex problems as well as keeping the discrimination in the conventional deep MIL framework.

%The optimization of the non-convex ${\cal F}\left( {{{\cal B}},w} \right)$ is to find a solution (network parameter) $w^*$, by
%\begin{equation}\label{eq_mil}
%    {w^*} = \mathop {\arg \min }\limits_w {{\cal F}\left( {{\cal B},w} \right)}.
%\end{equation}
%While directly optimizing Eq. \ref{eq_mil} causes local minimum solutions,

C-MIL roots in the traditional continuation method ~\cite{Allgower1990Numerical}, tracing a series of implicitly defined smoothed loss functions from a start point $(w^0 , 0)$ to a solution point $(w^* , 1)$, Fig.\ \ref{fig_motivation}(b), where $w^0$ is the solution of ${{\cal F}\left( {{\cal B},w, \lambda} \right)}$ when $\lambda=0$, and $w^*$ the solution when $\lambda=1$. Accordingly, we define a series of $\lambda$, $0=\lambda_0< \lambda_1< ...<\lambda_T=1$, and update Eq.\ \ref{eq_deep_MIL} to a continuation loss function, as
\begin{equation}\label{eq_cmil}
    \small
    \begin{split}
        {w^*} &= \mathop {\arg \min }\limits_w {{\cal F}\left( {{\cal B},w,\lambda} \right)} \\
              &= \mathop {\arg \min }\limits_{w_f,w_g}\sum\limits_i {{{\cal F}_f}\left( {{B_i},{B_{i,J(\lambda)}},{w_f}} \right) + {{\cal F}_g}\left( {{B_i},B_{i,J(\lambda)},{w_g}} \right)},
    \end{split}
\end{equation}
%{\color {blue} In this way, the non-convex loss function of the deep MIL framework is approximated with a seriers of smoothed loss functions. The continuation curve alleviates the non-convex problem and prevents the model fall into local minima too early. In the rest of this section, we will introduce how we define the continuation variable $\lambda$ for the instance selection and the detector estimation.}
where ${B_{i,J(\lambda)}} $ denotes the instance subset and
% $J(\lambda)$ the instance indexes in the subset, determined by parameter $\lambda$.
$J(\lambda)$ the index of ${B_{i,J(\lambda)}}$, determined by parameter $\lambda$. ${{\cal F}_f}\left( {{B_i},{B_{i,J(\lambda)}},{w_f}} \right)$ is the continuation loss function for instance selection, and ${{\cal F}_g}\left( {{B_i},{B_{i,J(\lambda)}},,{w_g}} \right)$  continuation loss function for detector estimation.

\begin{figure*}[!t]
    \begin{center}
       \includegraphics[width=0.95\linewidth]{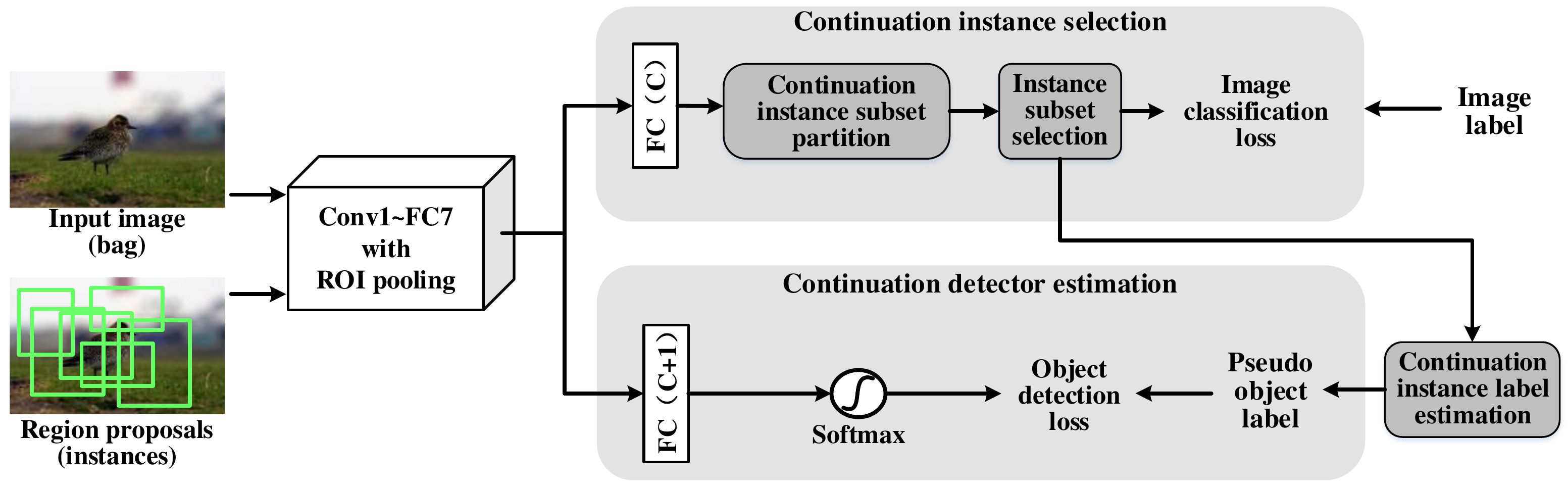}
    \end{center}
    \vspace{-0.3cm}
    \caption{The modules of continuation instance selection and continuation detector estimation are implemented atop a deep network for weakly supervised object detection. $C$ is the number of object categories. In the feed-forward procedure, C-MIL selects positive instances from subsets and uses them as pseudo-objects for detector estimation. In back-propagation, the instance selector and object detectors are jointly optimized with an SGD algorithm.}
    \label{fig-cmil-network}
    %\vspace{-0.5em}
\end{figure*}

\textbf{Continuation instance selection.}
%When learning the instance selector, we proposed to use a subset of instances instead of a single instance. This reduces the solution space and smooth the loss function of instance selection.
When learning the instance selector, a bag is partitioned into instance subsets, Fig.\ \ref{fig_instance_set}. In each subset object proposals are spatially related, $i.e.,$ overlapping with each other, and class related, $i.e.,$ having similar object class scores. The subsets are minimum sufficient cover to a bag (image) $B_i$, $i.e.,$ $\mathop \cup \limits_J B_{i,J} = {B_i}$ and $B_{i,J}^{} \cap B_{i,J'}^{} = \emptyset$ for $\forall J \ne J' $. All instances in a bag are sorted by their object scores $f\left( {{B_{i,j}},{w_f}} \right)$ and the following two steps are iteratively performed: %\textbf{\color{red}Say some words about spatially related and class-correlated.}

1) Construct an instance subset using the instance of highest object score while not belonging to any other instance subset. 2) Find the instances whose overlap with the highest scored instance $B_{i,j^*}$ are larger than or equal to $\lambda$, and then merge them into the subset.

When $\lambda=0$, bag $B_i$ is partitioned into a single subset which include all instances. When $\lambda=1$, a bag $B_i$ is partitioned into multiple subsets, each of which contains a single instance. The continuation of instance selection is performed from $\lambda=0$ to $\lambda=1$ with the loss function defined as
%The instance sub_sets  are minimum sufficient cover to a bag $B_i$ which satisfy $\mathop  \cup \limits_k S_{ik}^* = {B_i}$ and $S_{ik}^{} \cap S_{ik'}^{} = \emptyset$ for $\forall k \ne k' $. To partition the bag, all instances in the bag are sorted by their object scores $f\left( {{B_{i,j}},{w_f}} \right)$ and the following two steps are iteratively performed: 1) Construct an instance set using the instance of highest object score but not belonging to any instance set. 2) Find the instances whose overlap with the highest scored instance are larger than or equal to $\lambda$, and merge them into the instance set. The continuation instance mining loss is then defined as
\begin{equation}\label{eq_cmil_instance_miner}
    \small
    {{\cal F}_f}\left( {{B_i},{B_{i,J(\lambda)}},{w_f}} \right) = \max ( {0,1 - {y_i}\mathop {\max }\limits_{J(\lambda)} f( {B_{i,J(\lambda)},{w_f}} )}),
\end{equation}
where $f\left( {{B_{i,J(\lambda)}},{w_f}} \right)$, the score of instance subset $B_{i,J(\lambda)}$, is defined as
\begin{equation}\label{eq_instance_set}
    f\left( {{B_{i,J(\lambda)}},{w_f}} \right) = \frac {1}{\left| {B_{i,J(\lambda)}}\right|}\sum\nolimits_j {f\left( {{B_{i,j}},{w_f}} \right)},
\end{equation}
where $|{B_{i,J(\lambda)}}|$ denotes the number of instances in subset $B_{i,J(\lambda)}$ and $B_{i,j} \in B_{i,J(\lambda)}$.

During model learning, C-MIL equally utilizes all instances in subset $B_{i,J(\lambda)}$ to fine-tune the network parameters. As the instances are spatially overlapped and class related, C-MIL can collect object/parts for object extent activation, Fig. \ref{fig_instance_set}. When $\lambda=0$, each bag $B_i$ has a single subset that includes all instances. It is equal to change the term ${\max\nolimits_j f\left( {{B_{i,j}},{w_f}} \right)}$ of Eq.\ \ref{eq_deep_MIL_RegionMiner} to ${\sum\nolimits_j f\left( {{B_{i,j}},{w_f}} \right)}$ and then Eq.\ \ref{eq_cmil_instance_miner} becomes convex. When $\lambda=1$, a bag $B_i$ is partitioned into multiple subsets, each of which contains a single instance and thus Eq.\ \ref{eq_cmil_instance_miner} deteriorates to the original loss function, Eq.\ \ref{eq_deep_MIL_RegionMiner}. For $0< \lambda<1$, each bag $B_i$ has multiple subsets. According to Eq.\ \ref{eq_instance_set}, the score of an instance subset is equal to the average score of instances within that subset. The loss function Eq.\ \ref{eq_cmil_instance_miner} is therefore smoother than Eq.\ \ref{eq_deep_MIL_RegionMiner}, and then the loss function of CMIL defined by Eq.\ \ref{eq_cmil}, is smoother than that of MIL defined by Eq.\ \ref{eq_deep_MIL}. In other words, a series of smoothed loss functions are defined to alleviate the non-convexity problem of Eq.\ \ref{eq_deep_MIL_RegionMiner} and discover better solutions \cite{richter1983continuation, allgower1994numerical}, Fig.\ \ref{fig_motivation}(b).

%To explicitly show that the instance subset partition strategy introduces smoothness to Eq. \ref{eq_deep_MIL_RegionMiner}, we use the number of local optimal solutions as the \textit{smoothness metric} for a loss function. Given a bag of instance $B_i$, suppose the number of local optimal solutions is $K, K>1$. We then partition $B_i$ into instance subsets, with $L$ local optimal solutions.
%the set of local optimal solutions are $\{B_{i1^*},...,B_{ik^*},...,B_{iK^*}, B_{ik^*}\in B_i\}$, where $K$ is the number of local optimal solutions. By introducing the instance subset partition strategy, the local optimal solutions (instances) for Eq. \ref{eq_deep_MIL_RegionMiner} may be partitioned into that (instance subset) of Eq. \ref{eq_cmil_instance_miner}. Accordingly, we can infer that $K\ge L$. Specially, when $\lambda=0$, $L=1$ and therefore $K>L$, indicating that the instance subset partition strategy introduces smoothness for Eq. \ref{eq_deep_MIL_RegionMiner}. With the series of smoothed loss functions, the C-MIL is able to alleviate the non-convex problem of Eq.\ \ref{eq_deep_MIL_RegionMiner} and discovers better solutions \cite{richter1983continuation, allgower1994numerical}.
%\textbf{We must make it clear why subsets helps smoothing the objective/loss function.}

\textbf{Continuation detector estimation.}
During model learning, the subset $B_{i,J(\lambda)}$ of highest average score is selected for detector estimation. Considering that there is no bounding box annotation available, the instance selector is inaccurate and the selected subset might contain object parts or backgrounds. We further propose using a continuation strategy to estimate reliable instances and learn detectors.

%and the VOC detection metric \cite{fast-rcnn} is used to assign labels to instances. The instances whose overlap with the pseudo ground-truth larger than 0.5 are assigned with positive label; otherwise negative. However, in the early training stage, the instance miner is not able to find accurate instances. This will introduce lots of noise to the pseudo ground-truth and affect the model learning.
%The first strategy, as defined in Eq. \ref{eq_cmil}, is to use continuation variable $\lambda$ as a re regularization factor. During the early learning epoches, the selected instances are noisy so the weight of the detection loss is small. As the learning proceeds, the selected instances became accurate and the weight of the detection loss increases.
%As no bounding box annotation available, the instance selector might  will introduce false positive or false negative instances, especially when overlap with the pseudo ground-truth is near 0.5.
We propose to partition instances into positives and negatives with the continuation parameter $\lambda$.
Denote the learned instance subset as $B_{i,J(\lambda)^*}$ and the instance of highest score in $B_{i,J(\lambda)^*}$ as $B_{i,j^*}$. Instances in the bag are partitioned into positives or negatives according to their spatial relations, as
\begin{equation}\label{eq_label_assign}
    {y_{i,j}} = \left\{ {\begin{array}{*{20}{c}}
{+1,if\;IoU\left( {{B_{i,j}},B_{i,j^*}} \right) \ge 1 - \lambda /2}\\
{-1,if\;IoU\left( {{B_{i,j}},B_{i,j^*}} \right) < \lambda /2}\;\;\;\;\;\;\;
\end{array}}, \right.
\end{equation}
where $IoU$ calculates the Intersection of Union of two instances (bounding boxes). Eq.\ \ref{eq_label_assign} defines that instances whose IoU with $B_{i,j^*}$ greater than the threshold $1-\lambda/2$ are positives. Instances whose IoU with $B_{i,j^*}$ less than $\lambda/2$ are negatives. Instances whose IoU with $B_{i,j^*}$ falling into $[\lambda/2, 1-\lambda/2]$ are ignored.

During the learning procedure, along with the continuation parameter $\lambda$ changing from 0 to 1, the threshold $1-\lambda/2$ decreases from 1 to 0.5 and the threshold $\lambda/2$ increases from 0 to 0.5. According to Eq.\ \ref{eq_label_assign}, more and more instances are estimated as positives or negatives.
Based on these instances, the detector $g_z(B_{i,j}, w_g)$ is gradually estimated using the loss function defined as
% The detector is then gradually estimated, which implies a procedure of continuation optimization.
% In the procedure, instances with estimated labels are used to train the detector, as
\begin{equation}\label{eq_deep_MIL_ObjectDetector}
    % \small
    {{\cal F}_g}\left( {B_i,B_{i,J(\lambda)},{w_g}}\right) = -\sum\limits_z\sum\limits_j {{\delta_{z,y_{ij}}}\log g_z\left( {{B_{i,j}},{w_g}} \right)}.
\end{equation}
%is then gradually estimated, which implies the procedure of continuation optimization.
% where $y_{i,j^*}$ denotes the label of pseudo ground-truth $B_{i,j^*}$ and $y_{i,j}$ the label of $B_{i,j}$.

\subsection{Implementation}

C-MIL is implemented with an end-to-end deep neural network, with the continuation instance selection and continuation object estimation modules added atop of the FC layers, Fig.\ \ref{fig-cmil-network}.
In the training phrase, multiple instances, corresponding to region proposals, are first generated for each image using Selective Search method \cite{uijlings2013selective}. An ROI-pooling layer atop CONV5 and two fully connected layers are used for instance feature extraction. In the feed-forward procedure, C-MIL selects positive instances from subsets and uses them as pseudo-objects for detector estimation. In back-propagation, the instance selector and object detectors are jointly optimized with an SGD algorithm. With forward- and back-propagation procedures, network parameters are updated and the instance selector and object detectors are learned.
%We thus extend recurrent learning to accumulated recurrent learning, which accumulates different objects from both the object discovery and object localization branches. Doing so endows this approach with the capability to localize multiple objects in a single image but also provides the robustness necessary to process object appearance diversity by using multiple detectors.

%The MELMs are optimized with a recurrently learning algorithm, which uses forward propagation to select instances as pseudo objects, and back-propagation to optimize the network parameters with the gradient defined. The object probability of each instance is recurrently aggregated by multiplying by the object probability learned in the preceding iteration. In the detection phase, the learned object detectors, $i.e.$, the parameters for the soft-max and FC layers, are used to classify instances and localize objects.

The detection procedure involves instance feature extraction and instance classification  Fig.\ \ref{fig-cmil-network}. The learned detector computes object scores for all instances and Non-Maximum Suppression (NMS) is used to remove the overlapping instances.

%------------------------------------------------------------------------
\section{Experiments}
%------------------------------------------------------------------------
C-MIL was evaluated on the PASCAL VOC 2007 and PASCAL VOC 2012 datasets
%and MS-COCO 2014 dataset
using  mean average precision (mAP) \cite{everingham2010pascal} and correct localization (CorLoc) metrics \cite{deselaers2012Loc}, where Cor-Loc is the percentage of images for which the region of highest score has at least 0.5 interaction-over-union (IoU) with the ground-truth object region. In what follows, we first introduced the experimental settings, then analyzed the effect of the functions defined for the continuation parameter. The Stable Semantic Extremal Regions (SSERs) which appeared during the training procedure of C-MIL were also discussed. Finally, we reported the performance of C-MIL on WSOD and compared it with the state-of-the-art methods.

\subsection{Experimental Settings}
C-MIL was implemented based on the VGGF and VGG16 CNN model \cite{Simonyan2014Very} pre-trained on the ILSVRC 2012 dataset \cite{krizhevsky2012imagenet}.
%As the conventional object detection task \cite{Ren2016Weakly,Jie2017deepself},
We used Selective Search \cite{uijlings2013selective} to extract 2000 object proposals as instances for each image, and removed those whose width or height was less than 20 pixels.

The input images were re-sized into 5 scales \{480, 576, 688, 864, 1200\} with respect to the larger side (height or width). The scale of a training image was randomly selected and the image was randomly horizontal flipped. In this way, each test image was augmented into a total of 10 images \cite{Bilen2016Weakly,Tang2017OICR,Diba2017WCCN}. During learning, we employed the SGD algorithm with momentum 0.9, weight decay 5e-4, and batch size 1. The model iterated 20 epochs where the learning rate was 5e-3 for the first 10 epochs and 5e-4 for the last 10 epochs. During testing, the output scores of each instance from the 10 augmented images were averaged.

\begin{figure}[!t]
    \begin{center}
       \includegraphics[width=0.72\linewidth]{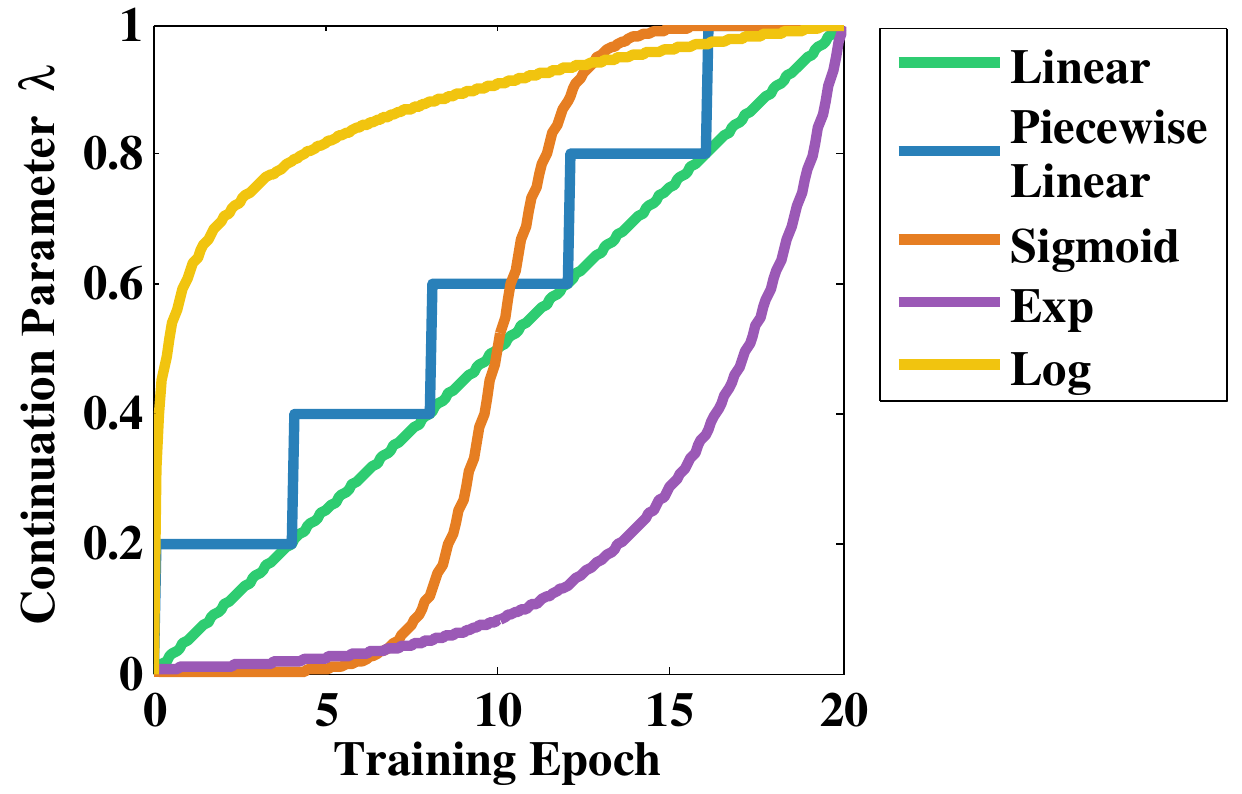}
    \end{center}
    \vspace{-0.3cm}
    \caption{Five functions defined to control the change of continuation parameter.}
    \label{fig_continuation_functions}
    %\vspace{-0.5em}
\end{figure}

\begin{figure}[!t]
    \begin{center}
       \includegraphics[width=0.95\linewidth]{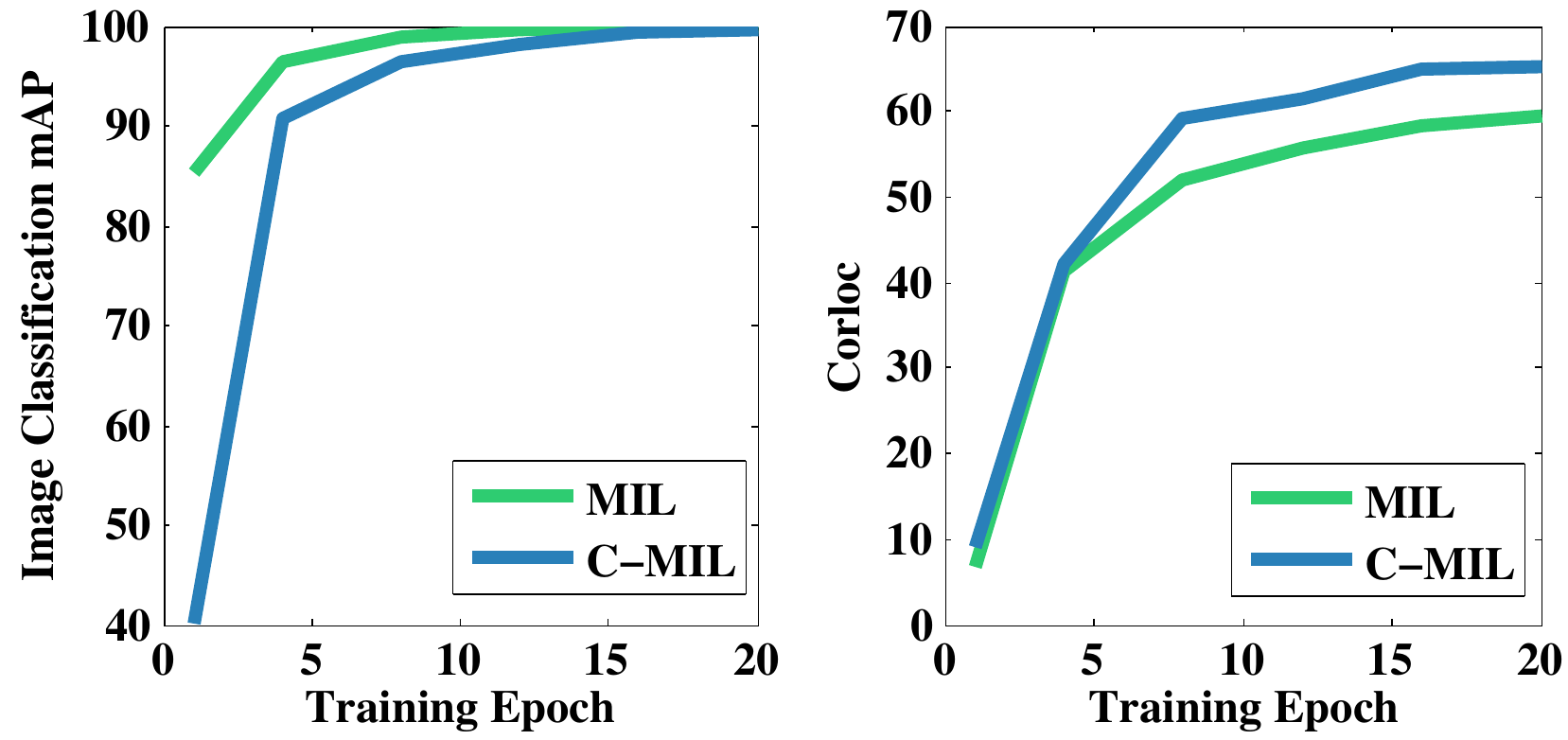}
    \end{center}
    \vspace{-0.3cm}
    \caption{Evolution of image classification and object localization performance during training.
    %In the early epochs, MIL achieves higher classification performance. In the later epochs, the classification performance of C-MIL catches up with that of MIL and the localization performance keeps higher than that of MIL.
    }
    \label{fig_performance_evolution}
    %\vspace{-0.5em}
\end{figure}

\begin{table}[!t]
\renewcommand{\arraystretch}{1.3}
\newcommand{\tabincell}[2]{\begin{tabular}{@{}#1@{}}#2\end{tabular}}
\caption{Comparison of five functions controlling the change of continuation parameter $\lambda$. Detection and localization performance (\%) on the VOC 2007 dataset with VGGF. }
\vspace{-0.2cm}
\label{table_effect_continuation_1}
\centering
\scriptsize
\begin{tabular}{
    @{}p{1.7cm}<{\centering}@{}|
    @{}p{3.5cm}<{\centering}@{}
    @{}p{1.2cm}<{\centering}@{}
    @{}p{1.2cm}<{\centering}@{}
    %@{}p{1cm}<{\centering}@{}
}
\hlinew{1.4pt}
\multirow{2}{*}{\tabincell{c}{Method}}
& Approaches /
& \multirow{2}{*}{\tabincell{c}{mAP}}
& \multirow{2}{*}{\tabincell{c}{CorLoc}} \\
& Continuation Functions &  &   \\
% state of the art methods
\hline
\multirow{1}{*}{\tabincell{c}{MIL}}
& ContextNet \cite{Kantorov2016ContextLocNet} &  36.0 & 55.0 \\
% & ContextNet (Ours baseline)   &  33.4 & 52.3 \\
\hline
\multirow{5}{*}{\tabincell{c}{C-MIL (Ours)}}
& Linear              &  37.9 & 58.9 \\
& Piecewise Linear    &  37.6 & 57.4 \\
& Sigmoid             &  38.3 & 58.4 \\
& Exp                 &  37.1 & 56.4 \\
& Log                 &  \textbf{40.7} & \textbf{59.5} \\
\hlinew{1.4pt}
\end{tabular}
\vspace{-0.2cm}
\end{table}

\begin{table}[!t]
\renewcommand{\arraystretch}{1.3}
\newcommand{\tabincell}[2]{\begin{tabular}{@{}#1@{}}#2\end{tabular}}
\caption{Ablation experimental results of C-MIL. Detection performance (\%) on the VOC 2007 dataset with VGGF.}
\vspace{-0.2cm}
\label{table_effect_continuation_2}
\centering
\scriptsize
\begin{tabular}{
    @{}p{2.2cm}<{\centering}@{}|
    @{}p{1.3cm}<{\centering}@{}
    @{}p{1.3cm}<{\centering}@{}
%    @{}p{1.5cm}<{\centering}@{}
    @{}p{1cm}<{\centering}@{}
}
\hlinew{1.4pt}
\multirow{2}{*}{\tabincell{c}{Method}}
%instance selector learning and detector learning
& Instance & Object   & \multirow{2}{*}{\tabincell{c}{mAP}}  \\
& Selector & Detector  &  \\
% state of the art methods
\hline
MIL \cite{Kantorov2016ContextLocNet} & - & - & 36.0 \\
\hline
\multirow{3}{*}{\tabincell{c}{C-MIL (Ours)}}
& \checkmark &            & 39.0 \\
&            & \checkmark & 37.4 \\
& \checkmark & \checkmark & 40.7 \\
\hlinew{1.4pt}
\end{tabular}
\vspace{-0.2cm}
\end{table}

\subsection{Continuation Method}
%We conduct ablation analyses of the proposed C-MIL, including the effect of the continuation functions for $\lambda$, the effect of continuation terms in C-MIL, and the discussion of SSER. All experiments in the ablation analyses are based on VOC 2007 benchmark using VGGF.
In this section, we investigated how to control the continuation parameter $\lambda$ and evaluated the effect on instance selection and detector estimation. All experiments are conducted on VOC 2007 benchmark.

\textbf{Continuation parameter $\lambda$.}
%The continuation functions control the {\color{blue} rate of change} of $\lambda$ during the training procedure.
To control the change rate of parameter $\lambda$ during training, five functions were evaluated, Fig.\ \ref{fig_continuation_functions}, and the results are shown in Table \ref{table_effect_continuation_1}. With continuation optimization the detection and localization performance respectively improved by 1.1\%$\sim$4.7\% and 1.4\%$\sim$4.5\%.

Table\ \ref{table_effect_continuation_1} shows that the ``Log'' function reported the best performance. With a ``Log'' function $\lambda$ increased quickly in the early training epochs while changed slowly in the late epochs, Fig.\ \ref{fig_continuation_functions}. This is consistent with the learning procedure: in the early training epochs, the instance subsets were large, and it required to dwindle them towards the positive instances; in the later epochs, the instance subsets tended to be stable and it required to focus on detector estimation.

%By comparing the five continuation functions, it can be seen that $\lambda$ should be small in the early epochs but soon be large in the later epochs (``Log''). In contrast, if the $\lambda$ keeps small for too many epochs (``Exp''), the learning of detector will be affected and the performance then drops. It shows that the conventional MIL often falls into local minima in the early training stage. C-MIL can alleviate this problem by introducing continuation optimization. In the rest of the training procedure, WSOD models should focus on the learning of detectors.

\begin{figure*}[!t]
    \begin{center}
       \includegraphics[width=0.97\linewidth]{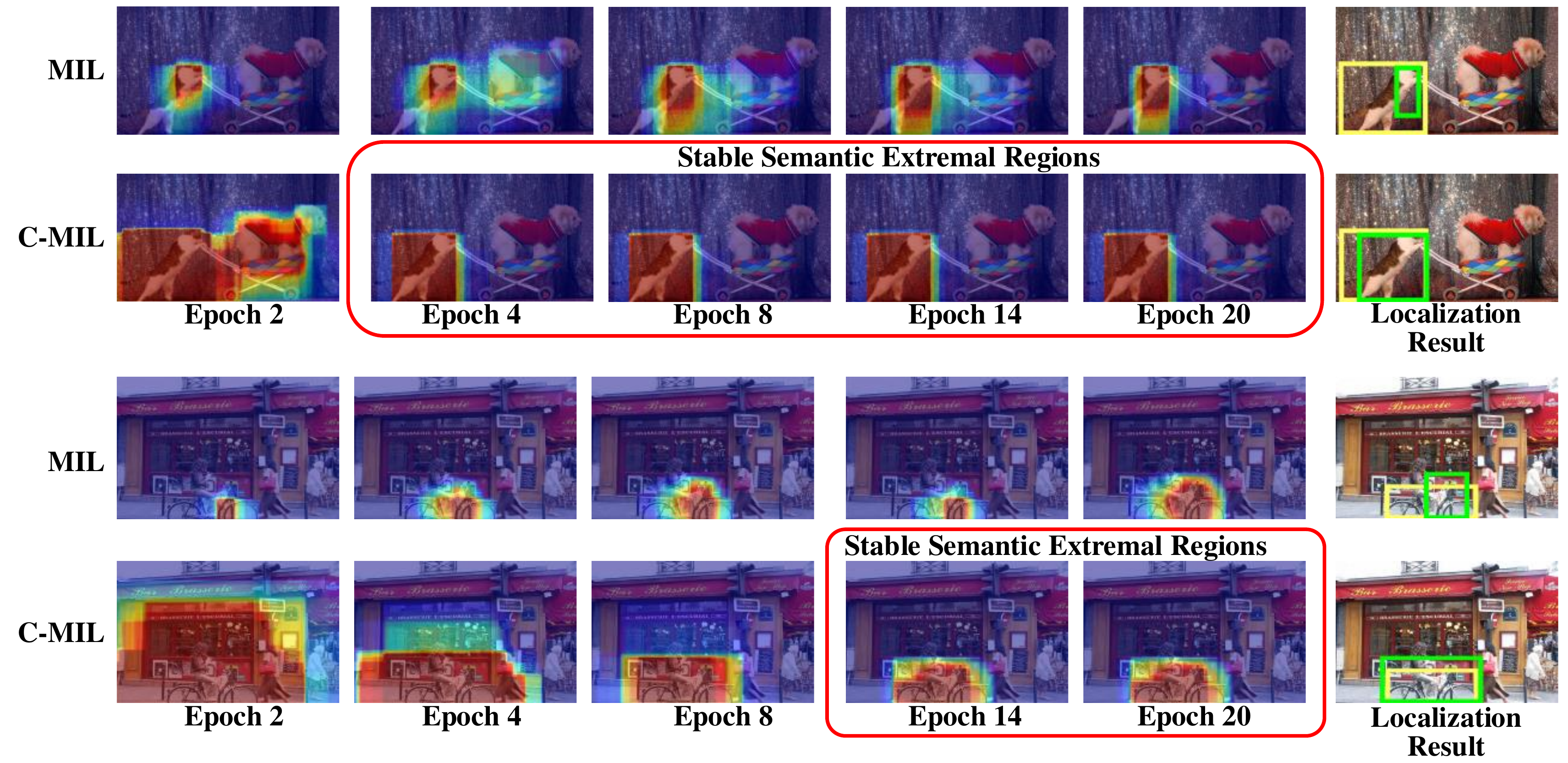}
    \end{center}
    \vspace{-0.6cm}
    \caption{Stable Semantic Extremal Regions (SSERs). MIL activated the discriminative regions for image classification but missed full object extent. C-MIL discovered SSERs indicating full object extent. The continuation parameter $\lambda$ of C-MIL increased from 0 to 1 along with the training procedure from epoch 0 to epoch 20. Yellow boxes and green boxes in the last column denote ground-truths and localization results, respectively. (Best viewed in color)}
    \label{fig_sser_results}
    %\vspace{-0.5em}
\end{figure*}

\textbf{Continuation optimization.}
Table\ \ref{table_effect_continuation_2} shows the ablation experimental results of continuation instance selection and continuation detector estimation. Compared with the baseline approach, introducing the continuation instance selection improved the performance by 3.0\% (39.0\% vs. 36.0\%); introducing the continuation of object estimation further improved the performance by 1.4\% (37.4\% vs. 36.0\%). Combining two modules aggregated the performance 4.7\% (40.7\% vs. 36.0\%), which clearly indicated the effectiveness of continuation optimization designed for C-MIL.

In Fig.\ \ref{fig_performance_evolution}, we visualized the evolution of the image classification and object localization during training. MIL achieved higher classification performance than C-MIL in the early training epochs. In the later epochs, the classification performance of C-MIL caught up with that of MIL and the localization performance kept higher than that of MIL. The reason lies in that MIL mainly optimized image classification without considering object localization. Therefore, it tended to discover regions which were discriminative for image classification but missed the object location. In contrast, C-MIL optimized both image classification and object localization by learning instance subsets, where object proposals are spatially related and class related. %C-MIL was able to activate full object extent.

\subsection{Stable Semantic Extremal Regions}

To understand the continuation optimization, we visualized the learned subsets/instances in different training epochs in Fig.\ \ref{fig_sser_results}. It can be seen that the instance subsets (activated regions) gradually dwindle with the increase of $\lambda$ from 0 to 1. In the early learning epochs, large subsets were defined to collect object/parts as many as possible. In the later learning epochs, the instance subsets stopped dwindling and tended to form stable activation regions around object boundaries. Such regions, referred to as Stable Semantic Extremal Region (SSERs), often turn out to be full object extent.

The emergence of SSERs indicated that C-MIL continuously suppressed backgrounds while activating object regions during learning.
%Since the semantics inside and outside the object boundary are not continuous, the instance subsets gradually dwindle to eliminate the backgrounds until it reaches the object boundary.
The procedure is somewhat similar to the process of extracting Maximally Stable Extremal Regions (MSERs) \cite{matas2004robust}. The difference lies in that the MSERs are defined for grey-level stable regions and extracted in an unsupervised manner while SSERs are defined for semantic stable regions and learned in a weakly supervised manner.

%In Fig.\ \ref{fig_sser_results}, we compared the activation regions of C-MIL and MIL. It can be seen that MIL activated the object part regions in the early epochs and thus soon get stack into local minima. In contrast, C-MIL activated the fully extent of objects. As the size of instance subset decreasing, C-MIL is able to discover the SSERs indicating the true object extent. %During network-fine-tuning, SSERs drives the network learning proper parameters to accurately localize objects.

\begin{table*}[!t]
\renewcommand{\arraystretch}{1.3}
\newcommand{\tabincell}[2]{\begin{tabular}{@{}#1@{}}#2\end{tabular}}
\caption{Detection performance (\%) on the VOC 2007 test set. Comparison of C-MIL to the state-of-the-arts.}
\vspace{-0.2cm}
\label{table_comp_voc07_state_of_the_art}
\centering
\scriptsize
\begin{tabular}{@{}p{0.82cm}<{\centering}|p{2.2cm}@{}
@{}m{0.62cm}<{\centering}@{}m{0.62cm}<{\centering}@{}m{0.62cm}<{\centering}@{}m{0.62cm}<{\centering}@{}m{0.77cm}<{\centering}@{}
@{}m{0.62cm}<{\centering}@{}m{0.57cm}<{\centering}@{}m{0.57cm}<{\centering}@{}m{0.62cm}<{\centering}@{}m{0.58cm}<{\centering}@{}
@{}m{0.62cm}<{\centering}@{}m{0.62cm}<{\centering}@{}m{0.62cm}<{\centering}@{}m{0.77cm}<{\centering}@{}m{0.77cm}<{\centering}@{}
@{}m{0.72cm}<{\centering}@{}m{0.77cm}<{\centering}@{}m{0.72cm}<{\centering}@{}m{0.62cm}<{\centering}@{}m{0.58cm}<{\centering}
|@{}m{0.8cm}<{\centering}@{}}
\hlinew{1.4pt}
Network & Method & aero & bike & bird & boat & bottle & bus & car & cat & chair & cow & table & dog & horse & mbike & person & plant & sheep & sofa & train & tv & mAP  \\
\hline
\multirow{9}{*}{\tabincell{c}{VGGF/\\AlexNet}}
% & MILinear \cite{Ren2016Weakly} &
% 41.3 & 39.7 & 22.1 &  9.5 &  3.9 & 41.0 & 45.0 & 19.1 &  1.0 & 34.0 &
% 16.0 & 21.3 & 32.5 & 43.4 & 21.9 & 19.7 & 21.5 & 22.3 & 36.0 & 18.0 & 25.4 \\
% & MF MIL \cite{cinbis2015multi-fold-J} &
% 39.3 & 43.0 & 28.8 & 20.4 &  8.0 & 45.5 & 47.9 & 22.1 &  8.4 & 33.5 &
% 23.6 & 29.2 & 38.5 & 47.9 & 20.3 & 20.0 & 35.8 & 30.8 & 41.0 & 20.1 & 30.2 \\
& PDA \cite{Li2016Weakly} &
49.7 & 33.6 & 30.8 & 19.9 & 13.0 & 40.5 & 54.3 & 37.4 & \textbf{14.8} & 39.8 &
9.4  & 28.8 & 38.1 & 49.8 & 14.5 & \textbf{24.0} & 27.1 & 12.1 & 42.3 & 39.7 & 31.0 \\
& LCL+Context \cite{wang2015LCL} &
48.9 & 42.3 & 26.1 & 11.3 & 11.9 & 41.3 & 40.9 & 34.7 & 10.8 & 34.7 &
18.8 & 34.4 & 35.4 & 52.7 & 19.1 & 17.4 & 35.9 & 33.3 & 34.8 & 46.5 & 31.6 \\
& WSDDN \cite{Bilen2016Weakly} &
42.9 & 56.0 & 32.0 & 17.6 & 10.2 & 61.8 & 50.2 & 29.0 &  3.8 & 36.2 &
18.5 & 31.1 & 45.8 & 54.5 & 10.2 & 15.4 & 36.3 & 45.2 & 50.1 & 43.8 & 34.5 \\
& ContextNet \cite{Kantorov2016ContextLocNet} &
\textbf{57.1} & 52.0 & 31.5 &  7.6 & 11.5 & 55.0 & 53.1 & 34.1 &  1.7 & 33.1 &
\textbf{49.2} & \textbf{42.0} & 47.3 & 56.6 & 15.3 & 12.8 & 24.8 & \textbf{48.9} & 44.4 & 47.8 & 36.3 \\
& WCCN \cite{Diba2017WCCN} &
43.9 & \textbf{57.6} & \textbf{34.9} & \textbf{21.3} & 14.7 & \textbf{64.7} & 52.8 & 34.2 &  6.5 & 41.2 &
20.5 & 33.8 & 47.6 & 56.8 & 12.7 & 18.8 & \textbf{39.6} & 46.9 & 52.9 & 45.1 & 37.3 \\
& OICR \cite{Tang2017OICR} &
53.1 & 57.1 & 32.4 & 12.3 & 15.8 & 58.2 & 56.7 & 39.6 &  0.9 & 44.8 &
39.9 & 31.0 & \textbf{54.0} & 62.4 &  4.5 & 20.6 & 39.2 & 38.1 & 48.9 & 48.6 & 37.9 \\
& MELM \cite{wan2018melm} &
56.4 & 54.7 & 30.9 & 21.1 & \textbf{17.3} & 52.8 & \textbf{60.0} & 36.1 &  3.9 & \textbf{47.8} &
35.5 & 28.9 & 30.9 & 61.0 &  5.8 & 22.8 & 38.8 & 39.6 & 42.1 & \textbf{54.8} & 38.4 \\
\cline{2-23}
& C-MIL (Ours) &
54.5 & 55.5 & 34.4 & 20.3 & 16.7 & 53.4 & 59.2 & \textbf{44.6} &  8.4 & 46.0 &
40.2 & 40.8 & 47.7 & \textbf{63.2} & \textbf{22.8} & 23.2 & 39.4 & 44.3 & \textbf{53.8} & 52.3 & \textbf{40.7} \\
\hlinew{1.4pt}
\multirow{8}{*}{VGG16}& WSDDN \cite{Bilen2016Weakly} &
39.4 & 50.1 & 31.5 & 16.3 & 12.6 & 64.5 & 42.8 & 42.6 & 10.1 & 35.7 &
24.9 & 38.2 & 34.4 & 55.6 &  9.4 & 14.7 & 30.2 & 40.7 & 54.7 & 46.9 & 34.8 \\
& PDA \cite{Li2016Weakly} &
54.5 & 47.4 & 41.3 & 20.8 & 17.7 & 51.9 & 63.5 & 46.1 & 21.8 & 57.1 &
22.1 & 34.4 & 50.5 & 61.8 & 16.2 & \textbf{29.9} & 40.7 & 15.9 & 55.3 & 40.2 & 39.5 \\
& OICR \cite{Tang2017OICR} &
58.0 & 62.4 & 31.1 & 19.4 & 13.0 & 65.1 & 62.2 & 28.4 & 24.8 & 44.7 &
30.6 & 25.3 & 37.8 & 65.5 & 15.7 & 24.1 & 41.7 & 46.9 & 64.3 & 62.6 & 41.2 \\
% & Self-Taught \cite{Jie2017deepself} &
% 52.2 & 47.1 & 35.0 & 26.7 & 15.4 & 61.3 & 66.0 & 54.3 &  3.0 & 53.6 &
% 24.7 & 43.6 & 48.4 & 65.8 &  6.6 & 18.8 & 51.9 & 43.6 & 53.6 & 62.4 & 41.7 \\
& WCCN \cite{Diba2017WCCN} &
49.5 & 60.6 & 38.6 & 29.2 & 16.2 & 70.8 & 56.9 & 42.5 & 10.9 & 44.1 &
29.9 & 42.2 & 47.9 & 64.1 & 13.8 & 23.5 & 45.9 & 54.1 & 60.8 & 54.5 & 42.8 \\
& TS$^2$C  \cite{Wei2018TS2C} &
59.3 & 57.5 & 43.7 & 27.3 & 13.5 & 63.9 & 61.7 & 59.9 & 24.1 & 46.9 &
36.7 & 45.6 & 39.9 & 62.6 & 10.3 & 23.6 & 41.7 & 52.4 & 58.7 & 56.6 & 44.3 \\
& WeakRPN  \cite{Tang2018WeakRPN} &
57.9 & \textbf{70.5} & 37.8 &  5.7 & \textbf{21.0} & 66.1 & \textbf{69.2} & 59.4 &  3.4 & 57.1 &
\textbf{57.3} & 35.2 & \textbf{64.2} & 68.6 & \textbf{32.8} & 28.6 & 50.8 & 49.5 & 41.1 & 30.0 & 45.3 \\
& MELM \cite{wan2018melm} &
55.6 & 66.9 & 34.2 & 29.1 & 16.4 & 68.8 & 68.1 & 43.0 & \textbf{25.0} & \textbf{65.6} &
45.3 & 53.2 & 49.6 & 68.6 &  2.0 & 25.4 & 52.5 & 56.8 & 62.1 & 57.1 & 47.3 \\
\cline{2-23}
& C-MIL (Ours) &
\textbf{62.5} & 58.4 & \textbf{49.5} & \textbf{32.1} & 19.8 & 70.5 & 66.1 & \textbf{63.4} & 20.0 & 60.5 &
52.9 & \textbf{53.5} & 57.4 & \textbf{68.9} & 8.4 & 24.6 & 51.8 & \textbf{58.7} & \textbf{66.7} & \textbf{63.5} & \textbf{50.5} \\
% & C-MIL &
% \textbf{62.5} & 58.4 & 49.5 & 32.1 & 19.8 & 70.5 & 66.1 & 63.4 & 20.0 & 60.5 &
% 52.9 & 53.5 & 57.4 & \textbf{68.9} & 8.4 & 24.6 & 51.8 & 58.7 & 66.7 & 63.5 & 50.5 \\
% & C-MIL+soft-nms &
% 62.3 & 61.1 & \textbf{52.8} & \textbf{34.1} & 20.8 & \textbf{71.2} & 67.4 & \textbf{65.0} & 20.5 & 62.1 &
% 55.1 & \textbf{56.6} & 59.6 & 68.5 & 9.7 & 25.4 & \textbf{54.8} & \textbf{59.6} & \textbf{69.5} & \textbf{64.2} & \textbf{52.0} \\
\hlinew{1.4pt}
& OICR-Ens. \cite{Tang2017OICR} &
\textbf{65.5} & 67.2 & 47.2 & 21.6 & 22.1 & 68.0 & 68.5 & 35.9 &  5.7 & 63.1 &
49.5 & 30.3 & 64.7 & 66.1 & 13.0 & 25.6 & 50.0 & 57.1 & 60.2 & 59.0 & 47.0 \\
FRCNN & TS$^2$C  \cite{Wei2018TS2C} &
-&-&-&-&-&-&-&-&-&-&-&-&-&-&-&-&-&-&-&-& 48.0 \\
Re-train& WeakRPN-Ens. \cite{Tang2018WeakRPN} &
63.0 & \textbf{69.7} & 40.8 & 11.6 & \textbf{27.7} & \textbf{70.5} & \textbf{74.1} & 58.5 & 10.0 & \textbf{66.7} &
\textbf{60.6} & 34.7 & \textbf{75.7} & \textbf{70.3} & \textbf{25.7} & \textbf{26.5} & \textbf{55.4} & 56.4 & 55.5 & 54.9 & 50.4 \\
\cline{2-23}
& C-MIL (Ours) &
61.8 & 60.9 & \textbf{56.2} & \textbf{28.9} & 18.9 & 68.2 & 69.6 & \textbf{71.4} & \textbf{18.5} & 64.3 &
57.2 & \textbf{66.9} & 65.9 & 65.7 & 13.8 & 22.9 & 54.1 & \textbf{61.9} & \textbf{68.2} & \textbf{66.1} & \textbf{53.1} \\
\hlinew{1.4pt}
\end{tabular}
\vspace{-0.2cm}
\end{table*}

\subsection{Performance}

Table \ref{table_comp_voc07_state_of_the_art} shows the performance of C-MIL and a comparison with the state-of-the-art methods on the PASCAL VOC 2007 dataset. It can be seen that C-MIL respectively achieved 40.7\% and 50.5\% with the VGGF and VGG16 models. With VGGF, C-MIL respectively outperformed the  WCCN \cite{Diba2017WCCN}, OICR \cite{Tang2017OICR}, and MELM \cite{wan2018melm} by 3.4\% (40.7\% vs. 37.3\%), 2.8\% (40.7\% vs. 37.9\%) and 2.3\% (40.7\% vs. 38.4\%). With VGG16, it respectively outperformed the WeakRPN \cite{Tang2018WeakRPN}, TS$^2$C \cite{Wei2018TS2C}, and MELM \cite{wan2018melm} by 6.2\% (50.5\% vs. 44.3\%), 5.2\% (50.5\% vs. 45.3\%), and 3.2\% (50.5\% vs. 47.3\%), which were large margins in terms of the challenging WSOD task.

We further re-trained an Fast-RCNN detector using the learned pseudo objects as ground-truth, and achieved 53.1\% mAP, as shown in Table \ref{table_comp_voc07_state_of_the_art}, which outperformed the state-of-the-art methods by 2.7\%$\sim$6.1\%. Specifically, the detection performance for ``aeroplane'' (+3.2\%), ``bird'' (+5.8\%), ``cat'' (+3.5\%), ``train'' (+4.5\%) significantly improved.
 %which shows {\color{red} the general effectiveness of C-MIL}.
% for weakly supervised object detection.

Table \ref{table_comp_voc12_state_of_the_art} shows the detection results of the proposed C-MIL and the state-of-the-art methods on the PASCAL VOC 2012 dataset with VGG16. For detection, C-MIL respectively outperformed the WeakRPN \cite{Tang2018WeakRPN}, TS$^2$C \cite{Wei2018TS2C}, and MELM \cite{wan2018melm} by 5.9\% (46.7\% vs. 40.8\%), 6.7\% (46.7\% vs. 40.0\%), and 4.3\% (46.7\% vs. 42.4\%).

We evaluated object localization performance of C-MIL and compared it with the state-of-the-art methods in Table \ref{table_comp_voc12_state_of_the_art} and Table \ref{table_comp_loc_state_of_the_art}. The used Correct Localization (CorLoc) metric \cite{deselaers2012Loc} is the percentage of images for which the region of highest object score has at least 0.5 interaction-over-union (IoU) with the ground-truth.
%This experiment was done on the $trainval$ set as the region selection exclusively worked in the training process.
It can be seen that C-MIL respectively outperformed the WeakRPN \cite{Tang2018WeakRPN} and TS$^2$C \cite{Wei2018TS2C} by 1.2\% (65.0\% vs. 63.8\%) and 4.0\% (65.0\% vs. 61.0\%) on VOC 2007,
%and respectively outperformed WeakRPN \cite{Tang2018WeakRPN} and TS$^2$C \cite{Wei2018TS2C}
and 3.0\% (67.4\% vs. 64.4\%) and 2.5\% (67.4\% vs. 64.9\%) on VOC 2012.

%To validate the applicability of C-MIL on large-scale object detection dataset, we conducted experiments on MS-COCO 2014 and reported the results in Table \ref{table_coco_2014}. It can be seen that C-MIL with a VGG16 network significantly outperformed the MIL-based approach (WSDDN \cite{Bilen2016Weakly}). With C-MIL, we set a solid baseline for weakly supervision object detection on the large-scale MS-COCO dataset.

\begin{table}[!t]
\renewcommand{\arraystretch}{1.3}
\newcommand{\tabincell}[2]{\begin{tabular}{@{}#1@{}}#2\end{tabular}}
\vspace{-0.2cm}
\caption{Detection and localization performance (\%) on the VOC 2012 dataset using VGG16. Comparison of C-MIL to the state-of-the-arts.}
\vspace{-0.2cm}
\label{table_comp_voc12_state_of_the_art}
\centering
\scriptsize
\begin{tabular}{@{}p{3.0cm}<{\centering}@{}|p{1.75cm}<{\centering}|@{}p{1.75cm}<{\centering}@{}}
\hlinew{1.4pt}
Method & mAP & CorLoc  \\
% state of the art methods
\hline
WCCN \cite{Diba2017WCCN}            & 37.9 & - \\
Self-Taught \cite{Jie2017deepself}  & 38.3 & 58.8 \\
OICR \cite{Tang2017OICR}            & 37.9 & 62.1 \\
TS$^2$C \cite{Wei2018TS2C}          & 40.0 & 64.4 \\
WeakRPN  \cite{Tang2018WeakRPN}     & 40.8 & 64.9 \\
MELM      \cite{wan2018melm}     & 42.4 & - \\
% our method
\cline{1-3}
C-MIL (Ours)                               & \textbf{46.7}  & \textbf{67.4} \\
\hlinew{1.4pt}
\end{tabular}
\vspace{-0.2cm}
\end{table}

\begin{table}[!t]
\renewcommand{\arraystretch}{1.3}
\newcommand{\tabincell}[2]{\begin{tabular}{@{}#1@{}}#2\end{tabular}}
\vspace{-0.2cm}
\caption{Localization performance (\%) on the VOC 2007 $trainval$ set. Comparison of C-MIL to the state-of-the-arts.}
\vspace{-0.2cm}
\label{table_comp_loc_state_of_the_art}
\centering
\scriptsize
\begin{tabular}{@{}p{2.0cm}<{\centering}@{}|p{3cm}<{\centering}|@{}p{1.5cm}<{\centering}@{}}
\hlinew{1.4pt}
CNN & Method & mAP  \\
% state of the art methods
\hline
\multirow{5}{*}{VGG16}
& WSDDN \cite{Bilen2016Weakly}      & 53.5 \\
& WCCN \cite{Diba2017WCCN}          & 56.7 \\
& OICR \cite{Tang2017OICR}          & 60.6 \\
& TS$^2$C \cite{Wei2018TS2C}        & 61.0 \\
& WeakRPN  \cite{Tang2018WeakRPN}   & 63.8 \\
% our method
\cline{2-3}
& C-MIL (Ours)                             & \textbf{65.0} \\
\hlinew{1.4pt}
\end{tabular}
\vspace{-0.2cm}
\end{table}

%\begin{table}[!t]
%\renewcommand{\arraystretch}{1.3}
%\newcommand{\tabincell}[2]{\begin{tabular}{@{}#1@{}}#2\end{tabular}}
% \vspace{-0.2cm}
%\caption{{\color{black}Detection performance (\%) on MS-COCO 2014. }}
%\vspace{-0.2cm}
%\label{table_coco_2014}
%\centering
%\scriptsize
%\begin{tabular}{
%p{2.4cm}<{\centering}|
%p{1.4cm}<{\centering}|
%p{1.4cm}<{\centering}|
%p{1.4cm}<{\centering}
%}
%\hlinew{1.4pt}
%Method & CNN & mAP@.5 & mAP@[.5,.95]   \\
%\hline
%MIL \cite{Bilen2016Weakly} & VGGF & 10.1 & 3.1  \\
%\hline
%\multirow{2}{*}{C-MIL (Ours)} & VGGF & 11.9 & 4.1  \\
%\cline{2-4}
% & VGG16 & \textbf{18.8} & \textbf{7.8}  \\
%\hlinew{1.4pt}
%\end{tabular}
%\vspace{-0.3cm}
%\end{table}

%\begin{figure*}[!t]
%    \begin{center}
%       \includegraphics[width=1\linewidth]{Fig-DetectionResults}
%    \end{center}
%    \vspace{-0.3cm}
%    \caption{Object detection examples on the PASCAL VOC 2007 datasets. Yellow bounding boxes denote ground-truth annotations and red boxes correct detection results. (Best viewed in color).}
%    \label{fig_detection_results}
%    %\vspace{-0.5em}
%\end{figure*}

\section{Conclusion}
We proposed an elegant and effective method, referred to as C-MIL, for weakly supervised object detection. C-MIL targets alleviating the non-convexity problem of multiple instance learning using a series of smoothed loss functions. These functions were defined by introducing a parametric strategy for instance subset partition and evaluating the training loss according to these subsets in a deep learning framework. C-MIL significantly improved performance of weakly supervised object detection and weakly supervised object localization, in striking contrast with state-of-the-art approaches. The underlying reality is that the continuation optimization combined the deep feature learning first collects object/object parts to activate true object extent and then discovers Stable Semantic Extremal Regions (SSERs) for object localization. This provides a fresh insight for the weakly supervised object detection problem.

\textbf{Acknowledgments.} The authors are very grateful to the support by NSFC grant 61836012, 61771447, and 61671427, and Beijing Municipal Science and Technology Commission grant Z181100008918014.

{\small
\bibliographystyle{ieee_fullname}
\bibliography{refs}

\begin{thebibliography}{10}\itemsep=-1pt

\bibitem{krizhevsky2012imagenet}
Krizhevsky Alex, Sutskever Ilya, and Hinton~Geoffrey E.
\newblock Imagenet classification with deep convolutional neural networks.
\newblock In {\em Adv. in Neural Inf. Process. Syst. ({NIPS})}, pages
  1097--1105, 2012.

\bibitem{Allgower1990Numerical}
Eugene~L. Allgower and Kurt Georg.
\newblock {\em Numerical Continuation Methods}.
\newblock 1990.

\bibitem{allgower1994numerical}
Eugene~L Allgower, Kurt Georg, and R Hettich.
\newblock Numerical continuation methods. an introduction.
\newblock {\em Jahresbericht der Deutschen Mathematiker Vereinigung},
  96(1):26--26, 1994.

\bibitem{beck2012smoothing}
Amir Beck and Marc Teboulle.
\newblock Smoothing and first order methods: A unified framework.
\newblock {\em SIAM Journal on Optimization}, 22(2):557--580, 2012.

\bibitem{bengio2009curriculum}
Yoshua Bengio, J{\'e}r{\^o}me Louradour, Ronan Collobert, and Jason Weston.
\newblock Curriculum learning.
\newblock In {\em Proc. 26st Int. Conf. Mach. Learn. ({ICML})}, pages 41--48.
  ACM, 2009.

\bibitem{bilen2014}
Hakan Bilen, Marco Pedersoli, and Tinne Tuytelaars.
\newblock Weakly supervised object detection with posterior regularization.
\newblock In {\em Brit. Mach. Vis. Conf. ({BMVC})}, pages 1997--2005, 2014.

\bibitem{bilen2015}
Hakan Bilen, Marco Pedersoli, and Tinne Tuytelaars.
\newblock Weakly supervised object detection with convex clustering.
\newblock In {\em Proc. IEEE Int. Conf. Comput. Vis. Pattern Recognit.
  ({CVPR})}, pages 1081--1089, 2015.

\bibitem{Bilen2016Weakly}
Hakan Bilen and Andrea Vedaldi.
\newblock Weakly supervised deep detection networks.
\newblock In {\em Proc. IEEE Int. Conf. Comput. Vis. Pattern Recognit.
  ({CVPR})}, pages 2846--2854, 2016.

\bibitem{chaudhari2017entropy}
Pratik Chaudhari, Anna Choromanska, Stefano Soatto, Yann LeCun, Carlo Baldassi,
  Christian Borgs, Jennifer Chayes, Levent Sagun, and Riccardo Zecchina.
\newblock Entropy-sgd: Biasing gradient descent into wide valleys.
\newblock In {\em Int. Conf. Learn. Repres.}, 2017.

\bibitem{chen2012smoothing}
Xiaojun Chen.
\newblock Smoothing methods for nonsmooth, nonconvex minimization.
\newblock {\em Mathematical programming}, 134(1):71--99, 2012.

\bibitem{wang2015LCL}
Wang Chong, Huang Kaiqi, Ren Weiqiang, Zhang Junge, and Maybank Steve.
\newblock Large-scale weakly supervised object localization via latent category
  learning.
\newblock {\em {IEEE} Trans. Image Process.}, 24(4):1371--1385, 2015.

\bibitem{wang2014LCL}
Wang Chong, Ren Weiqiang, Huang Kaiqi, and Tan Tieniu.
\newblock Weakly supervised object localization with latent category learning.
\newblock In {\em Proc. Europ. Conf. Comput. Vis. ({ECCV})}, pages 431--445,
  2014.

\bibitem{clevert2015fast}
Djork-Arn{\'e} Clevert, Thomas Unterthiner, and Sepp Hochreiter.
\newblock Fast and accurate deep network learning by exponential linear units
  (elus).
\newblock {\em arXiv preprint arXiv:1511.07289}, 2015.

\bibitem{Diba2017WCCN}
Ali Diba, Vivek Sharma, Ali Pazandeh, Hamed Pirsiavash, and Luc Van~Gool.
\newblock Weakly supervised cascaded convolutional networks.
\newblock In {\em Proc. IEEE Int. Conf. Comput. Vis. Pattern Recognit.
  ({CVPR})}, pages 5131--5139, 2017.

\bibitem{Li2016Weakly}
Li Dong, Huang~Jia Bin, Li Yali, Wang Shengjin, and Yang~Ming Hsuan.
\newblock Weakly supervised object localization with progressive domain
  adaptation.
\newblock In {\em Proc. IEEE Int. Conf. Comput. Vis. Pattern Recognit.
  ({CVPR})}, pages 3512--3520, 2016.

\bibitem{gao2018cwsl}
Mingfei Gao, Ang Li, Ruichi Yu, Vlad~I Morariu, and Larry~S Davis.
\newblock C-wsl: Count-guided weakly supervised localization.
\newblock 2018.

\bibitem{girshick2015fast-rcnn}
Ross Girshick.
\newblock Fast r-cnn.
\newblock In {\em Proc. IEEE Int. Conf. Comput. Vis. Pattern Recognit.
  ({CVPR})}, pages 1440--1448, 2015.

\bibitem{cinbis2014multi-fold-C}
Cinbis~Ramazan Gokberk, Verbeek Jakob, and Schmid Cordelia.
\newblock Multi-fold mil training for weakly supervised object lcalization.
\newblock In {\em Proc. IEEE Int. Conf. Comput. Vis. Pattern Recognit.
  Workshop}, pages 2409--2416, 2014.

\bibitem{cinbis2015multi-fold-J}
Cinbis~Ramazan Gokberk, Verbeek Jakob, and Schmid Cordelia.
\newblock Weakly supervised object localization with multi-fold multiple
  instance learning.
\newblock {\em {IEEE} Trans. Pattern Anal. Mach. Intell.}, 39(1):189--203,
  2016.

\bibitem{Gulcehre2017Mollifying}
Caglar Gulcehre, Marcin Moczulski, Francesco Visin, and Yoshua Bengio.
\newblock Mollifying networks.
\newblock In {\em Int. Conf. Learn. Repres.}, 2017.

\bibitem{Jie2017deepself}
Zequn Jie, Yunchao Wei, Xiaojie Jin, Jiashi Feng, and Wei Liu.
\newblock Deep self-taught learning for weakly supervised object localization.
\newblock In {\em Proc. IEEE Int. Conf. Comput. Vis. Pattern Recognit.
  ({CVPR})}, pages 4294--4302, 2017.

\bibitem{Kantorov2016ContextLocNet}
Vadim Kantorov, Maxime Oquab, Minsu Cho, and Ivan Laptev.
\newblock Contextlocnet: Context-aware deep network models for weakly
  supervised localization.
\newblock In {\em Proc. Europ. Conf. Comput. Vis. ({ECCV})}, pages 350--365,
  2016.

\bibitem{Simonyan2014Very}
Simonyan Karen and Zisserman Andrew.
\newblock Very deep convolutional networks for large-scale image recognition.
\newblock In {\em ICLR}, 2015.

\bibitem{Zitnick2014Edge}
Zitnick~C. Lawrence and Dollár Piotr.
\newblock Edge boxes: Locating object proposals from edges.
\newblock In {\em Proc. Europ. Conf. Comput. Vis. ({ECCV})}, pages 391--405,
  2014.

\bibitem{everingham2010pascal}
Everingham Mark, Van~Gool Luc, Williams~Christopher KI, Winn John, and
  Zisserman Andrew.
\newblock The pascal visual object classes (voc) challenge.
\newblock {\em Int. J. Comput. Vis}, 88(2):303--338, 2010.

\bibitem{matas2004robust}
Jiri Matas, Ondrej Chum, Martin Urban, and Tom{\'a}s Pajdla.
\newblock Robust wide-baseline stereo from maximally stable extremal regions.
\newblock {\em Image and vision computing}, 22(10):761--767, 2004.

\bibitem{song2014pattern}
Song~Hyun Oh, Lee~Yong Jae, Jegelka Stefanie, and Darrell Trevor.
\newblock Weakly supervised discovery of visual pattern configurations.
\newblock In {\em Adv. in Neural Inf. Process. Syst. ({NIPS})}, pages
  1637--1645, 2014.

\bibitem{Song2014On}
Song~Hyun Oh, Girshick Ross, Jegelka Stefanie, Mairal Julien, Harchaoui Zaid,
  and Darrell Trevor.
\newblock On learning to localize objects with minimal supervision.
\newblock In {\em Proc. 31st Int. Conf. Mach. Learn. ({ICML})}, pages
  1611--1619, 2014.

\bibitem{Siva2011Weakly}
Siva Parthipan and Xiang Tao.
\newblock Weakly supervised object detector learning with model drift
  detection.
\newblock In {\em Proc. IEEE Int. Conf. Comput. Vis. ({ICCV})}, pages 343--350,
  2011.

\bibitem{ren2015faster-rcnn}
Shaoqing Ren, Kaiming He, Ross Girshick, and Jian Sun.
\newblock Faster r-cnn: Towards real-time object detection with region proposal
  networks.
\newblock In {\em Adv. in Neural Inf. Process. Syst. ({NIPS})}, pages 91--99,
  2015.

\bibitem{richter1983continuation}
Stephen~L Richter and Raymond~A Decarlo.
\newblock Continuation methods: Theory and applications.
\newblock {\em {IEEE} Trans. on Sys. Man and Cyber.}, (4):459--464, 1983.

\bibitem{uijlings2013selective}
Uijlings~Jasper RR, Van de Sande Koen~EA, Gevers Theo, and Smeulders~Arnold WM.
\newblock Selective search for object recognition.
\newblock {\em Int. J. Comput. Vis}, 104(2):154--171, 2013.

\bibitem{andrews2002MIL}
Andrews Stuart, Tsochantaridis Ioannis, and Hofmann Thomas.
\newblock Support vector machines for multiple-instance learning.
\newblock In {\em Adv. in Neural Inf. Process. Syst. ({NIPS})}, pages 561--568,
  2002.

\bibitem{Tang2017OICR}
Peng Tang, Xinggang Wang, Xiang Bai, and Wenyu Liu.
\newblock Multiple instance detection network with online instance classifier
  refinement.
\newblock In {\em Proc. IEEE Int. Conf. Comput. Vis. Pattern Recognit.
  ({CVPR})}, pages 3059--3067, 2017.

\bibitem{Tang2018WeakRPN}
Peng Tang, Xinggang Wang, Angtian Wang, Yongluan Yan, Wenyu Liu, Junzhou Huang,
  and Alan Yuille.
\newblock Weakly supervised region proposal network and object detection.
\newblock In {\em Proc. Europ. Conf. Comput. Vis. ({ECCV})}, pages 352--368,
  2018.

\bibitem{deselaers2012Loc}
Deselaers Thomas, Alexe Bogdan, and Ferrari Vittorio.
\newblock Weakly supervised localization and learning with generic knowledge.
\newblock {\em Int. J. Comput. Vis}, 100(3):275--293, 2012.

\bibitem{wan2018melm}
Fang Wan, Pengxu Wei, Jianbin Jiao, Zhenjun Han, and Qixiang Ye.
\newblock Min-entropy latent model for weakly supervised object detection.
\newblock In {\em Proc. IEEE Int. Conf. Comput. Vis. Pattern Recognit.
  ({CVPR})}, pages 1297--1306, 2018.

\bibitem{Wei2018TS2C}
Yunchao Wei, Zhiqiang Shen, Bowen Cheng, Honghui Shi, Jinjun Xiong, Jiashi
  Feng, and Thomas Huang.
\newblock Ts2c:tight box mining with surrounding segmentation context for
  weakly supervised object detection.
\newblock In {\em Proc. Europ. Conf. Comput. Vis. ({ECCV})}, pages 434--450,
  2018.

\bibitem{Ren2016Weakly}
Ren Weiqiang, Huang Kaiqi, Tao Dacheng, and Tan Tieniu.
\newblock Weakly supervised large scale object localization with multiple
  instance learning and bag splitting.
\newblock {\em {IEEE} Trans. Pattern Anal. Mach. Intell.}, 38(2):405--416,
  2016.

\bibitem{Ye2017SelLearning}
Qixiang Ye, Tianliang Zhang, Qiang Qiu, Baochang Zhang, Jie Chen, and Guillermo
  Sapiro.
\newblock Self-learning scene-specific pedestrian detectors using a progressive
  latent model.
\newblock In {\em Proc. IEEE Int. Conf. Comput. Vis. Pattern Recognit.
  ({CVPR})}, pages 2057--2066, 2017.

\bibitem{zheng2015improving}
Hao Zheng, Zhanlei Yang, Wenju Liu, Jizhong Liang, and Yanpeng Li.
\newblock Improving deep neural networks using softplus units.
\newblock In {\em Proc. IEEE Int. Joint Conf. Neural Networks ({IJCNN})}, pages
  1--4, 2015.

\end{thebibliography}
}

\end{document}